\documentclass[letterpaper, 10 pt, conference]{ieeeconf}  % Comment this line out if you need a4paper

\IEEEoverridecommandlockouts                              % This command is only needed if 
\usepackage[linesnumbered,ruled]{algorithm2e}
\usepackage{float}
\usepackage{subfigure}
\usepackage{graphics} % for pdf, bitmapped graphics files
\usepackage{epsfig} % for postscript graphics files
\usepackage{times} % assumes new font selection scheme installed
\usepackage{amsmath} % assumes amsmath package installed
\usepackage{amssymb}  % assumes amsmath package installed
\usepackage{color}
\usepackage[table]{xcolor}
\usepackage[T1]{fontenc}
\usepackage{bm}
\usepackage{cite}
\usepackage{hyperref}
\usepackage{multirow}
\hypersetup{
	colorlinks=true,
	linkcolor=blue,
	filecolor=blue,      
	urlcolor=blue,
	citecolor=cyan,
}

\title{\LARGE \bf Keyfilter-Aware Real-Time UAV Object Tracking}

\author{Yiming Li$^{1}$, Changhong Fu$^{1,*}$, Ziyuan Huang$^{2}$, Yinqiang Zhang$^{3}$, and Jia Pan$^{4}$% <-this % stops a space
\thanks{$^{1}$Yiming Li and Changhong Fu are with the School of Mechanical Engineering, Tongji University, 201804 Shanghai, China. \tt\small changhongfu@tongji.edu.cn}%
\thanks{$^{2}$Ziyuan Huang is with the Advanced Robotics Centre, National University of Singapore, Singapore. \tt \small ziyuan.huang@u.nus.edu}%
\thanks{$^{3}$Yinqiang Zhang is with the Department of Mechanical Engineering, Technical University of Munich, Munich,  Germany. \tt\small yinqiang.zhang@tum.de} 
\thanks{$^{4}$Jia Pan is with the Computer Science Department, The University of Hong Kong, Hong Kong, China.
        {\tt\small panjia1983@gmail.com}}%
}

\begin{document}

\maketitle
\thispagestyle{empty}
\pagestyle{empty}

%%%%%%%%%%%%%%%%%%%%%%%%%%%%%%%%%%%%%%%%%%%%%%%%%%%%%%%%%%%%%%%%
%%%%%%%%%%%%%%%%%%%%% Section 0: abstract %%%%%%%%%%%%%%%%%%%%%%
%%%%%%%%%%%%%%%%%%%%%%%%%%%%%%%%%%%%%%%%%%%%%%%%%%%%%%%%%%%%%%%%
\begin{abstract}
Correlation filter-based tracking has been widely applied in unmanned aerial vehicle (UAV) with high efficiency. However, it has two imperfections, i.e., boundary effect and filter corruption. Several methods enlarging the search area can mitigate boundary effect, yet introducing undesired background distraction. Existing frame-by-frame context learning strategies for repressing background distraction nevertheless lower the tracking speed. Inspired by keyframe-based simultaneous localization and mapping, keyfilter is proposed in visual tracking for the first time, in order to handle the above issues efficiently and effectively. Keyfilters generated by periodically selected keyframes learn the context intermittently and are used to restrain the learning of filters, so that 1) context awareness can be transmitted to all the filters via keyfilter restriction, and 2) filter corruption can be repressed. Compared to the state-of-the-art results, our tracker performs better on two challenging benchmarks, with enough speed for UAV real-time applications.
\end{abstract}

%%%%%%%%%%%%%%%%%%%%%%%%%%%%%%%%%%%%%%%%%%%%%%%%%%%%%%%%%%%%%%%%
%%%%%%%%%%%%%%%%%%%%% Section 1: INTRODUCTION %%%%%%%%%%%%%%%%%%
%%%%%%%%%%%%%%%%%%%%%%%%%%%%%%%%%%%%%%%%%%%%%%%%%%%%%%%%%%%%%%%%
\section{INTRODUCTION}
Combined with extensibility, autonomy, and maneuverability of unmanned aerial vehicle (UAV), visual object tracking has considerable applications in UAV, e.g., person tracing~\cite{Cheng2017IROS}, autonomous landing~\cite{Fu2014ICUAS}, aerial photography~\cite{Bonatti2019IROS}, and aircraft tracking~\cite{Fu2014ICRA}. Notwithstanding some progress, UAV tracking remains onerous because of the complex background, frequent appearance variation caused by UAV motion, full/partial occlusion, deformation, as well as illumination changes. Besides, computationally intractable trackers are not deployable onboard UAVs because of the harsh calculation resources and limited power capacity.
     
Recently, the framework of discriminative correlation filter (DCF)~\cite{Henriques2015TPAMI}, aiming to discriminate the foreground from the background via a correlation filter (CF), is widely adopted in UAV object tracking. The speed is hugely raised because of its utilization of the circulant matrices' property to carry out the otherwise cumbersome calculation in the frequency domain rather than spatial one. Yet the circulant artificial samples used to train the filter hamper the filter's discriminative ability. This problem is called boundary effect because the artificial non-real samples have periodical splicing at the boundary. Several approaches~\cite{Danelljan2015ICCV,Danelljan2016ECCV,Danelljan2017CVPR,Fu2019IROS,Luke2017CVPR,Galoogahi2015CVPR,Galoogahi2017ICCV,Huang2019ICCV} expand the search area for alleviating boundary effects,  but the enlargement has introduced more context noise, distracting the detection phase especially in situations of similar objects around. 

In literature, the context-aware framework~\cite{Mueller2017CVPR} is proposed to reduce the context distraction through response repression of the context patches. However, the frame-by-frame context learning is extremely redundant, because the capture frequency of drone camera is generally smaller than the frequency of context variation, e.g., the interval time between two consecutive time in a 30 frame per second (FPS) video is 0.03 second, but generally the context appearance in aerial view remains unchanged for a certain time far more than 0.03 second. In addition, the learned single filter without restriction is prone to corruption due to the omnipresent appearance variations in the aerial scenarios. 
%%%%%%%%%%%%%%%%%%%
\begin{figure}[t]
	\centering
	\includegraphics[width=0.495\textwidth]{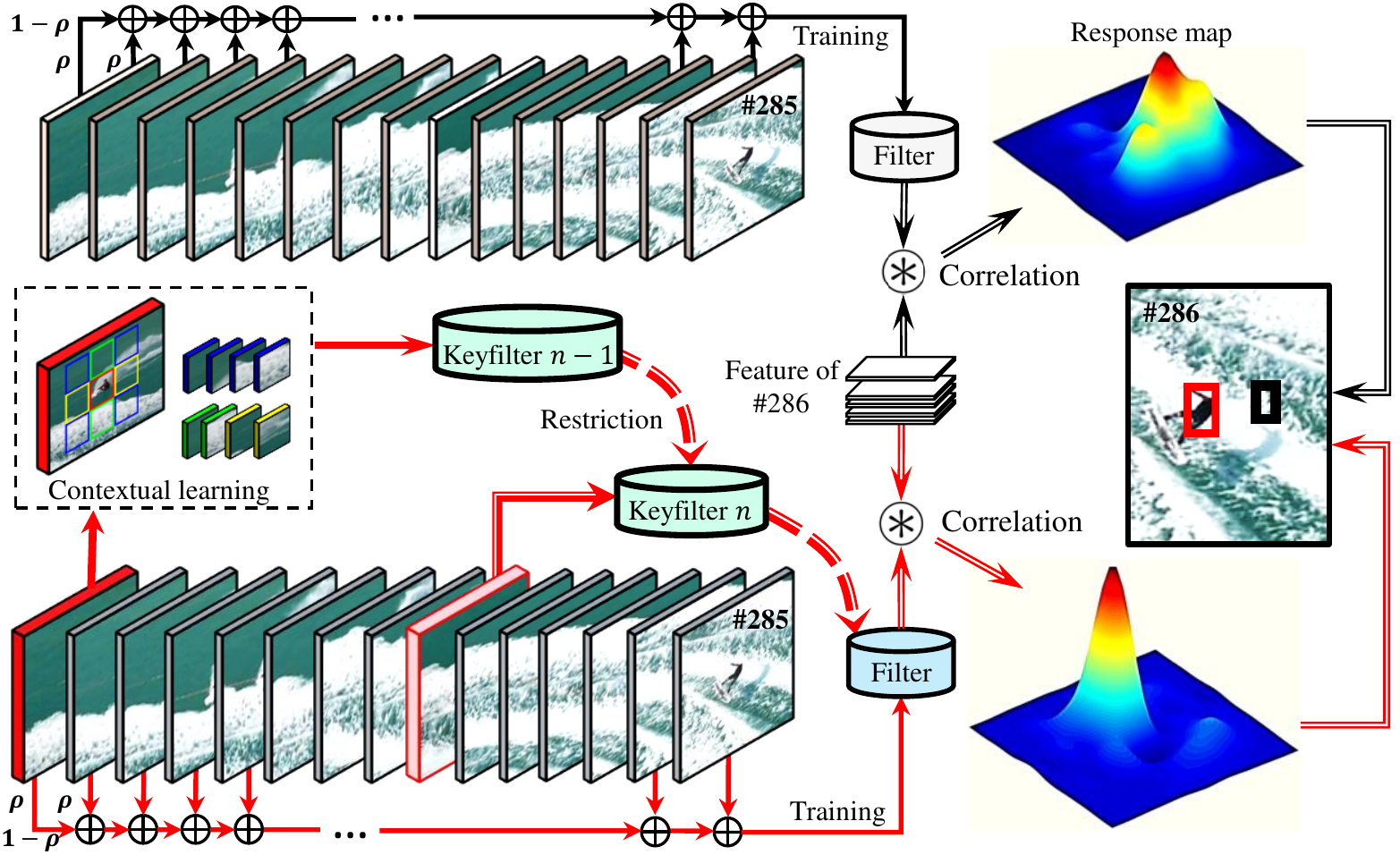}
	\caption{Comparison between response maps of our tracker and baseline. Red frames are served as keyframes generating keyfilters. Keyfilters carry out context learning intermittently and influence the current filter training for mitigating filter corruption. Feature of current frame is correlated with the filter trained in the last frame, producing a response map. Red and black rectangles denote respectively the results from KAOT and baseline.}
	\label{fig:scene}
\end{figure}
%%%%%%%%%%%%%%%%%%

In this work, inspired by keyframe-based simultaneous localization and mapping (SLAM)~\cite{Artal2015Trans}, the keyframe technique is used to raise the tracking performance efficiently and effectively. The contributions of this work are two-fold:
\begin{itemize}
	\item A novel application of the keyfilter in UAV visual object tracking is presented. Keyfilters generated at a certain frequency learn the context intermittently and enforce temporal restriction. Through the restriction, the filter corruption in the time span is alleviated and  context noise is efficiently suppressed.
	\item  Extensive experiments on 193 challenging UAV image sequences have shown that the \textbf{k}eyfilter-\textbf{a}ware \textbf{o}bject \textbf{t}racker, i.e., KAOT,  has competent performance compared with the state-of-the-art tracking approaches based on DCF and deep neural network (DNN).
\end{itemize}
%%%%%%%%%%%%%%%%%%%%%%%%%%%%%%%%%%%%%%%%%%%%%%%%%%%%%%%%%%%%%%%%
%%%%%%%%%%%%%%%%%%%%% Section 2: RELATED WORK %%%%%%%%%%%%%%%%%%
%%%%%%%%%%%%%%%%%%%%%%%%%%%%%%%%%%%%%%%%%%%%%%%%%%%%%%%%%%%%%%%%
\section{RELATED WORKS}\label{sec:RELATEDWORK}
\subsection{Discriminative correlation filter}
In recent years, the framework of discriminative correlation filter (DCF)~\cite{Henriques2015TPAMI} has broadly aroused research interest due to its remarkable efficiency. Yet classic CF-based trackers~\cite{Bertinetto2016TPAMI,Yang2015ECCVW,Ma2015CVPR} have limited performance due to the lack of negative samples, i.e., the circulant artificial samples created to train the CF hugely reduce its discriminative power. One solution to this problem is spatial penalization to punish the filter value at the boundary~\cite{Danelljan2015ICCV,Danelljan2016ECCV,Danelljan2017CVPR,Fu2019IROS,Luke2017CVPR}. Another solution is cropping both the background and target to use negative samples in the real word instead of synthetic samples~\cite{Galoogahi2015CVPR,Galoogahi2017ICCV,Huang2019ICCV}. However, the aforementioned approaches are prone to introduce context distraction because of enlarging search area, especially in the scenarios of similar object around.
\subsection{Prior work to context noise and filter corruption}
In literature, M. Mueller et al.~\cite{Mueller2017CVPR} proposed to repress the response of context patches, i.e., the features extracted from surrounding context are directly fed into classic DCF framework and their desired responses are suppressed as zero. The context distraction is thus effectively repressed, consequently the discriminative ability of the filter is enhanced. Nevertheless, the frame-by-frame context learning is effective but not efficient, and its redundancy can be significantly reduced. Another problem of classic DCF trackers is that the learned single filter is commonly subjected to corruption because of the frequent appearance variation. Online passive-aggressive learning is incorporated into the DCF framework~\cite{Li2018CVPR} to mitigate the corruption. Compared to~\cite{Li2018CVPR}, the presented keyfilter performs better in both precision and speed. 
\subsection{Tracking by deep neural network}
Recently, deep neural network has contributed a lot to the development of computer vision. For visual tracking, some deep trackers~\cite{wang2015transferring,Nam2016CVPR,Wang2015ICCV} fine-tuning the deep network online for high precision yet run too slow (around 1 fps on a high-end GPU) to use in practice. Other methods like deep reinforcement learning \cite{Yun2017CVPR}, unsupervised learning~\cite{Wang2019CVPR}, continues operator \cite{Danelljan2017CVPR}, end-to-end learning~\cite{Valmadre2017CVPR} and deep feature representation~\cite{Ma2015ICCV} have also increased the tracking accuracy.  Among them, incorporating lightweight deep features into online learned DCF framework has exhibited competitive performance both in precision and efficiency. 
\subsection{Tracking for unmanned aerial vehicle}
Mechanical vibration, motion blur, limited computation capacity and rapid movement have made UAV tracking an extremely demanding task. In literature, the presented UAV-tailored tracking methods generally have lower robustness and accuracy~\cite{Fu2014ICRA,2016YinTrans,Yuan2015ICUAS,Martinez2013JINT}. In light of offline training on the large-scale image datasets, deep feature for robust representation can improve performance significantly, yet the speed of existing deep-feature based trackers mostly run slow even on a high-end GPU~\cite{Fu2019IROS}. This work aims to improve the speed and accuracy for the deep feature-based DCF framework for real-time UAV applications.  

%%%%%%%%%%%%%%%%%%%%%%%%%%%%%%%%%%%%%%%%%%%%%%%%%%%%%%%%%%%%%%%%
%%%%%%%%%%%%%%%%%%%%% Section 3: Prior     Works %%%%%%%%%%%%%%%
%%%%%%%%%%%%%%%%%%%%%%%%%%%%%%%%%%%%%%%%%%%%%%%%%%%%%%%%%%%%%%%%
\section{REVIEW OF BACKGROUND-AWARE CORRELATION FILTER}\label{sec:PriorWorks}
The objective function of background-aware correlation filters (BACF) \cite{Galoogahi2017ICCV} is as follows:
%%%%%%%%%%%%%%%%%%%%%%%%%%%%%%%%
\begin{equation}\label{BACFobjective}
\small
\mathcal { E } ( \mathbf { w } ) = \frac { 1 } { 2 } \Vert \mathbf { y } - \sum _ { d = 1 } ^ { D } \mathbf { Bx }_0 ^ { d } \star \mathbf { w } ^ { d } \Vert _2^2 +\frac { \lambda } { 2 } \sum _ { d = 1 } ^ { D } \Vert \mathbf { w } ^ { d } \Vert _2^2 \ ,
\end{equation}
%%%%%%%%%%%%%%%%%%%%%%%%%%%%%%%%
\noindent where $\mathbf{y} \in \mathbb{R}^M$, $\mathbf{x}^d\in \mathbb{R}^N$ and $\mathbf{w}^d\in \mathbb{R}^M$  denote the desired response, the $d$th one of $D$ feature channels and correlation filter respectively. $\lambda$ is a regularization parameter and $\mathcal{E}(\mathbf{w})$ refers to an error  between the desired response $\mathbf{y}$ and the actual one. $\star$ is the spatial correlation operator. The main idea of BACF is to utilize a cropping matrix $\mathbf{B} \in \mathbb{R}^{M\times N}$ to extract real negative samples. However, more background distraction is introduced because of the enlargement.
%%%%%%%%%%%%%%%%%%%%%%%%%%%%%%%%
%%%%%%%%%%%

%%%%%%%%%%%
%%%%%%%%%%%%%%%%%%%%%%%%%%%%%%%%%%%%%%%%%%%%%%%%%%%%%%%%%%%%%%%%
%%%%%%%%%%%%%%%%%%%%% Section 4: PROPOSED METHOD %%%%%%%%%%%%%%%
%%%%%%%%%%%%%%%%%%%%%%%%%%%%%%%%%%%%%%%%%%%%%%%%%%%%%%%%%%%%%%%%
\section{KEYFILTER-AWARE OBJECT TRACKER}\label{sec:KAOT}
\begin{figure*}[!t]
	\centering
	\includegraphics[width= 0.991\textwidth]{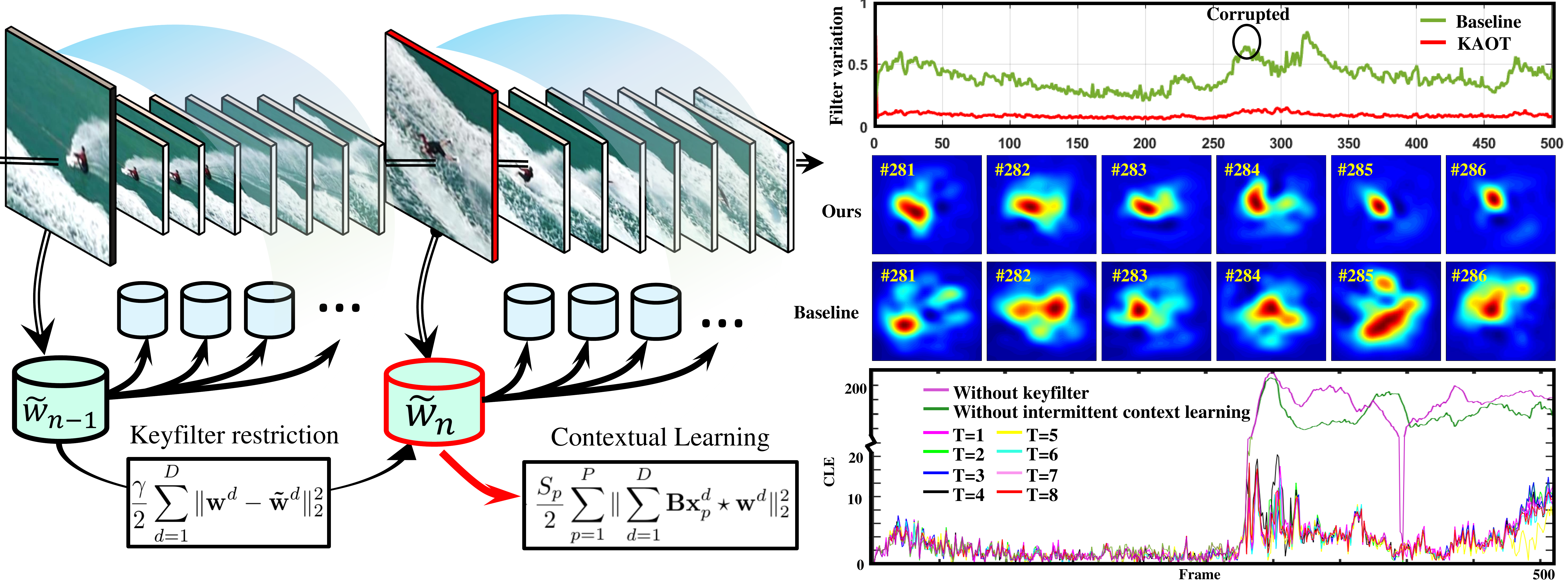}
	\caption{\textbf{Illustration of advantages of KAOT.} With the keyfilter restriction, the filter corruption is mitigated, as shown on the top right. With the context learning, the distraction is reduced, as shown in the response maps from frame 281 to 286. Set the keyfilter update period T as 1-8 frames (learns the context every 2 - 16 frames), and the object is tracked successfully in all eight trackers, while FPS (frame per second) is raised to 15.2 from 9.8, lowering the redundancy of context learning significantly. In addition, trackers lacking of the keyfilter restriction or the context learning both lose the target.
	}
	\label{fig:mainstructure}
\end{figure*}

Inspired by the keyframe technique used in SLAM, the keyfilter is firstly proposed in visual tracking to boost accuracy and efficiency, as illustrated in Fig.~\ref{fig:mainstructure}. The objective function of KAOT tracker is written as follows:
%%%%%%%%%%%%%%%%%%%%%%%%%%%%%%%%
\begin{equation}\label{eq:objective}
\small
\begin{aligned}
\mathcal { E } ( \mathbf { w } ) &= \frac { 1 } { 2 } \Vert \mathbf { y } - \sum _ { d = 1 } ^ { D } \mathbf { Bx }_0 ^ { d } \star \mathbf { w } ^ { d } \Vert _2^2 +\frac { \lambda } { 2 } \sum _ { d = 1 } ^ { D } \Vert \mathbf { w } ^ { d } \Vert _2^2 \\
&+\frac {S_p} { 2 }\sum _ { p = 1 } ^ { P } \Vert \sum _ { d = 1 } ^ { D }  \mathbf { Bx }_p ^ { d } \star \mathbf { w } ^ { d }\Vert _2^2 +\frac { \gamma } { 2 }\sum _ { d = 1 } ^ { D }\Vert \mathbf { w }^{ d }-\mathbf {\tilde{w}  }^{d} \Vert _2^2 
\end{aligned}\ ,
\end{equation}
%%%%%%%%%%%%%%%%%%%%%%%%%%%%%%%%

\noindent where the third term is response repression of context patches (their desired responses are zero), and $S_p$ is the score of $p$th patch to measure the necessity of penalization (introduced in~\ref{subsec:scoring}). $\mathbf{w}^d\in \mathbb{R}^M$ and $\mathbf{\tilde{w} }^d\in \mathbb{R}^M$  are the current filter and keyfilter, respectively. $\gamma$ is the penalty parameter of the gap between $\mathbf{w}^d$ and $\mathbf{\tilde{w} }^d$. To improve the calculation speed, Eq.~(\ref{eq:objective}) is calculated in the frequency domain:
%%%%%%%%%%%%%%%%%%
\begin{small}
\begin{equation}
\label{matrix}
\begin{aligned}
\mathcal{E}(\mathbf{w}, \hat{\mathbf{g}}) = \frac{1}{2} \Vert  &\hat{\mathbf{X}} \hat{\mathbf{g}} -\hat{\mathbf{Y}} \Vert _2^2 + \frac{\lambda}{2}\Vert \mathbf{\mathbf{w}} \Vert _2^2 + \frac{\gamma}{2}\Vert \mathbf { w } - \mathbf {\tilde{w} } \Vert _2^2 \\ 
&s.t. \quad \hat{\mathbf{g}} = \sqrt{N}(\mathbf{I}_D \otimes \mathbf{F} \mathbf{B}^{\top} )\mathbf{w}
\end{aligned}  \ ,
\end{equation}
\end{small}
%%%%%%%%%%%%%%%%%%

\noindent where $\otimes$ is the Kronecker product and $\mathbf{I}_D \in \mathbb{R}^{D \times D}$ is an identity matrix. $ \hat{}$ denotes the discrete Fourier transform with orthogonal matrix $\mathbf{ F }$. $\hat{\mathbf { X }}^{T} = \left[ \begin{array} { c } { \hat {\mathbf { X }} _ { 0 } },  { \hat {S_1\mathbf { X }} _ { 1 } },\cdots,  {  S_p\hat {\mathbf { X }} _ { P } } \end{array} \right] $, $\hat {\mathbf { Y }} = \left[ \begin{array} { c } { \hat{\mathbf { y } }},  \mathbf{ 0 },\cdots,  \mathbf{0} \end{array} \right] $, and $\hat{\mathbf{X}}_p \in \mathbb{C}^{N\times DN} $($p=0,1,...,P$), $\hat{\mathbf{g}} \in \mathbb{C}^{DN\times 1}$ and $\mathbf{w} \in \mathbb{R}^{DM\times 1}$ are respectively defined as $\hat{\mathbf{X}} = [ diag(\hat{\mathbf{x}}^1)^\top, \cdots, diag(\hat{\mathbf{x}}^D)^\top]$, $\hat{\mathbf{g}} = [\hat{\mathbf{g}}^{1\top} , \cdots , \hat{\mathbf{g}}^{D\top}]^\top$, $\mathbf{\tilde{w}} = [\mathbf{\tilde{w}}^{1\top} , \cdots , \mathbf{\tilde{w}}^{D\top}]^\top$  and $\mathbf{w} = [\mathbf{w}^{1\top} , \cdots , \mathbf{w}^{D\top}]^\top$.
%%%%%%%%%%%%%%%%%%%%%%%%%%%%%%%%%%%%%%%%%%%%%%%%%%%%%%%%%%%%%
\subsection{Optimization algorithm}\label{Optimization}
Equation (\ref{matrix}) can be optimized via alternating direction method of multipliers (ADMM)~\cite{Stephen2010FTML}. The Augmented Lagrangian form of Eq.~(\ref{matrix}) is:
%%%%%%%%%%%%%
\begin{equation}\label{equ:L(w,g,zeta)}
\small
\begin{split}
\mathcal{L}(\mathbf{w}, \hat{\mathbf{g}}, \hat{\bm\zeta})
& = \frac{1}{2} \Vert  \hat{\mathbf{X}} \hat{\mathbf{g}} -\hat{\mathbf{Y}} \Vert _2^2 + \frac{\lambda}{2}\Vert \mathbf{\mathbf{w}} \Vert _2^2 + \frac{\gamma}{2}\Vert \mathbf { w } - \mathbf {\tilde{w}} \Vert _2^2 \\
& \quad + \hat{\bm\zeta}^\top \big( \hat{\mathbf{g}} - \sqrt{N}(\mathbf{I}_D \otimes \mathbf{F} \mathbf{B}^{\top})\mathbf{w} \big)\\
& \quad + \frac{\mu}{2} \Vert \hat{\mathbf{g}} - \sqrt{N}(\mathbf{I}_D \otimes \mathbf{F} \mathbf{B}^{\top})\mathbf{w}\Vert_2^2 
\end{split} \ ,
\end{equation}
%%%%%%%%%%%%%
\noindent where  $\hat{\bm\zeta} \in \mathbb{C}^{DN\times 1}$ is the Lagrangian vector in the frequency domain and $\mu$ is a penalty parameter. Two subproblems $\hat{\mathbf{g}}^*$ and $\mathbf{w}^*$ are solved alternatively.
%%%%%%%%%%%%%%%%%%%%%%%%%%%%%%%%%%%%%%%%

\noindent$\bullet$ \textbf{Subproblem $\mathbf{w}^*$ (filter in the spatial domain):}
%%%%%%%%%%%%%%%%
\begin{small}
\begin{equation}\label{subproblem_w}
\begin{aligned} 
\mathbf { w } ^ { * } = & \arg \min _ { \mathbf { w } } \left\{   \frac{\lambda}{2}\Vert \mathbf{\mathbf{w}} \Vert _2^2 + \frac{\gamma}{2}\Vert \mathbf { w } - \mathbf {\tilde{w}} \Vert _2^2 \right.\\
& \quad + \hat{\bm\zeta}^\top \big( \hat{\mathbf{g}} - \sqrt{N}(\mathbf{I}_D \otimes \mathbf{F} \mathbf{B}^{\top})\mathbf{w} \big)\\
& \left.\quad + \frac{\mu}{2} \Vert \hat{\mathbf{g}} - \sqrt{N}(\mathbf{I}_D \otimes \mathbf{F} \mathbf{B}^{\top})\mathbf{w}\Vert_2^2 \right\} \\ 
& = \left( \mu + \frac { \lambda+\gamma } {  { N } } \right) ^ { - 1 } ( \mu \mathbf { g } + \zeta + \frac { \gamma } { N }\mathbf{\tilde{w}}) 
\end{aligned} \ .
\end{equation}
\end{small}

%\noindent where $\mathbf{ g}$ and $\mathbf{\zeta }$ are $\mathbf{g} =  \frac { 1 } { \sqrt { N } } \left(\mathbf { I } _ { D } \otimes \mathbf { FB } ^ { \top } \right) \hat { \mathbf { g } }$ and $\mathbf { \zeta } = \frac { 1 } { \sqrt { N } } \left(  \mathbf { I } _ { D } \otimes \mathbf { FB } ^ { \top } \right) \hat { \mathbf { \zeta } }$, respectively.
%%%%%%%%%%%%%%%

\noindent$\bullet$ \textbf{Subproblem $\hat{\mathbf{g}}^*$ (filter in the frequency domain):}
%%%%%%%%%%%%%%%%
\begin{small}
\begin{equation}
\begin{aligned} 
\hat{\mathbf{g}}^* = & \arg \min _ {\hat{\mathbf{g}}} \left\{   \frac{1}{2} \Vert  \hat{\mathbf{X}} \hat{\mathbf{g}} -\hat{\mathbf{Y}} \Vert _2^2 \right.\\
& \quad + \hat{\bm\zeta}^\top \big( \hat{\mathbf{g}} - \sqrt{N}(\mathbf{I}_D \otimes \mathbf{F} \mathbf{B}^{\top})\mathbf{w} \big)\\
& \left.\quad + \frac{\mu}{2} \Vert \hat{\mathbf{g}} - \sqrt{N}(\mathbf{I}_D \otimes \mathbf{F} \mathbf{B}^{\top})\mathbf{w}\Vert_2^2 \right\}
\end{aligned}\ .
\end{equation}
\end{small}
%%%%%%%%%%%%%%%%

$\hat {\mathbf{y}}(n)$  only depends on  $\hat{\mathbf{x}}(n)=\left[\hat {\mathbf{x}}^1(n),\hat {\mathbf{x}}^2(n),...,\hat {\mathbf{x}}^D(n)\right]^{\top}$
and $\hat { \mathbf { g } } ( n ) = \left[ \operatorname { conj } \left( \hat { \mathbf { g } } ^ { 1 } ( n ) \right) , \ldots , \operatorname { conj } \left( \hat { \mathbf { g } } ^ { D } ( n ) \right) \right] ^ { \top }$. Hence, solving equation  for $\hat{\mathbf{g}}^*$ can be identically written as $N$ separate functions  $\hat{\mathbf{g }}(n)$ ($n = [1, ..., N ]$):
%%%%%%%%%%%%%%%%%%%%%%%%%%%%%%%%%%%%
\begin{equation}
\label{gnn}
\small
\begin{aligned} 
\hat { \mathbf { g } } ( n ) ^ { * } &=  \arg \min _ { \hat { \mathbf { g } } ( n ) } \left\{ \frac { 1 } { 2 } \Vert \hat { \mathbf { y } } ( n ) - \hat { \mathbf { x }}_0  ( n ) ^ { \top } \hat { \mathbf { g } } ( n ) \Vert_2^2 \right. \\ 
&+\frac{1}{2} \sum _ { p = 1 } ^ { P }\Vert S_p\hat { \mathbf { x }}_p  ( n ) ^ { \top } \hat { \mathbf { g } } ( n ) \Vert_2^2+ \hat { \zeta } ( n ) ^ { \top } ( \hat { \mathbf { g } } ( n ) - \hat { \mathbf { w } } ( n ) ) \\ 
&\left. + \frac { \mu } { 2 } \| \hat { \mathbf { g } } ( n ) - \hat { \mathbf { w } } ( n ) \| _ { 2 } ^ { 2 } \right\} 
\end{aligned}\ ,
\end{equation}
%%%%%%%%%%%%%%%%%%%%%%%%%%%%%%%%%%%%

\noindent where $\hat { \mathbf { w } } ( n ) = \left[ \hat { \mathbf { w } } ^ { 1 } ( n ) , \ldots , \hat { \mathbf { w } } ^ { D } ( n ) \right]$ and $\hat { \mathbf { w } } ^ { d } = \sqrt { D } \mathbf { F } \mathbf { P } ^ { \top } \mathbf { w } ^ { d }$. 
The solution to each sub-subproblem is:
%%%%%%%%%%%%%%%%%%%%%%%%%%%%%%%%%%%%
\begin{small}
\begin{equation}\label{equ:subprobg}
\begin{aligned} \hat { \mathbf { g } } ( n ) ^ { * } = ( \sum_{p=0}^{P} &S_p^2\hat { \mathbf { x }}_p  ( n ) \hat { \mathbf { x }} _p ( n ) ^ { \top } 
 + \mu \mathbf { I } _ { D } ) ^ { - 1 } \\
&( \hat { \mathbf { y } } ( n ) \hat { \mathbf { x }_0 } ( n ) -  \hat { \zeta } ( n ) + \mu \hat { \mathbf { w } } ( n ) ) 
\end{aligned}\ .
\end{equation}
\end{small}
\begin{figure*}[!t]
	\centering
	\subfigure{
		\begin{minipage}[t]{0.47\linewidth}
			\centering
			\includegraphics[width=3.1in]{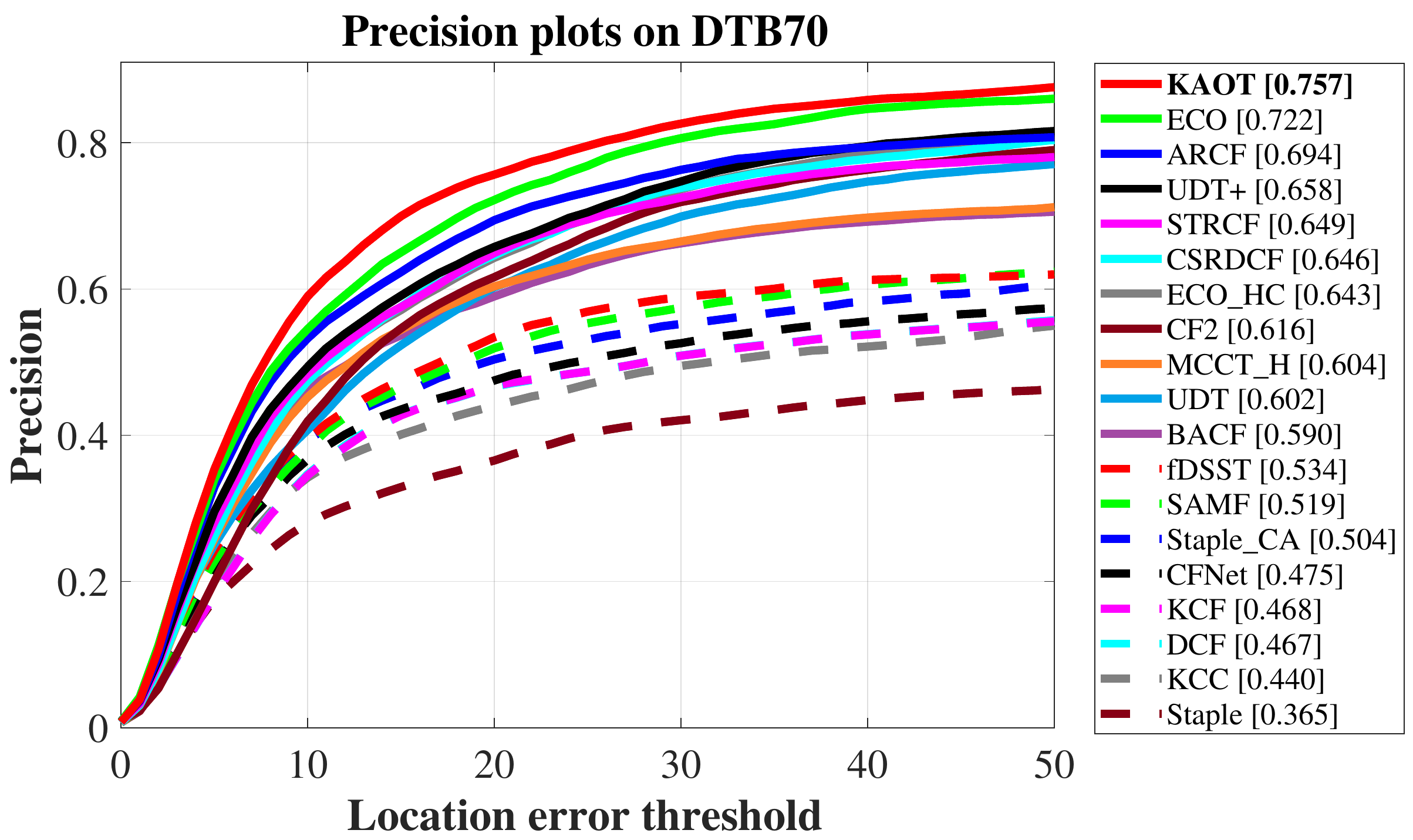}
			%\caption{fig1}
		\end{minipage}%
	}%
	\subfigure{
		\begin{minipage}[t]{0.47\linewidth}
			\centering
			\includegraphics[width=3.1in]{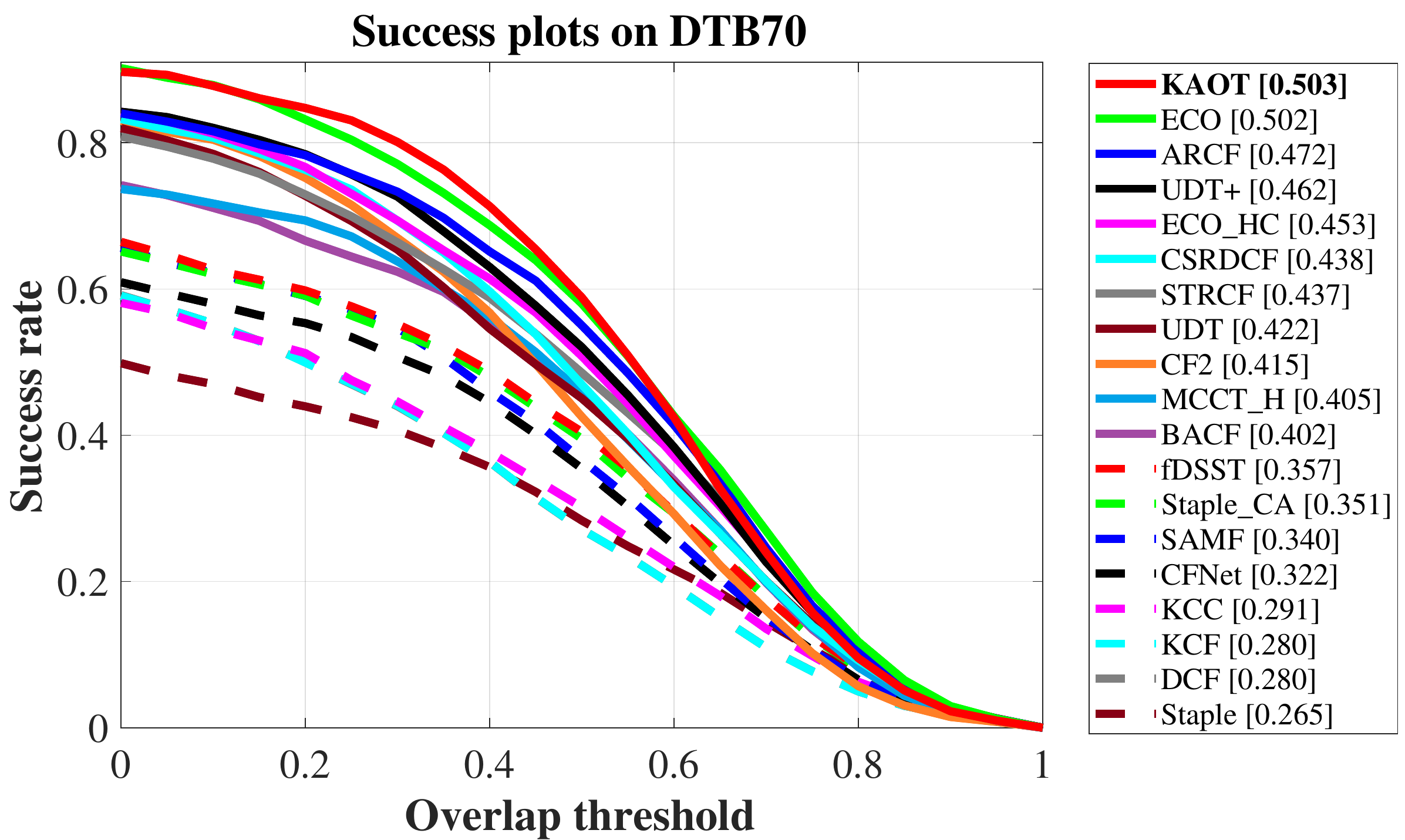}
			%\caption{fig2}
		\end{minipage}%
	}%
	
	\subfigure{
		\begin{minipage}[t]{0.47\linewidth}
			\centering
			\includegraphics[width=3.1in]{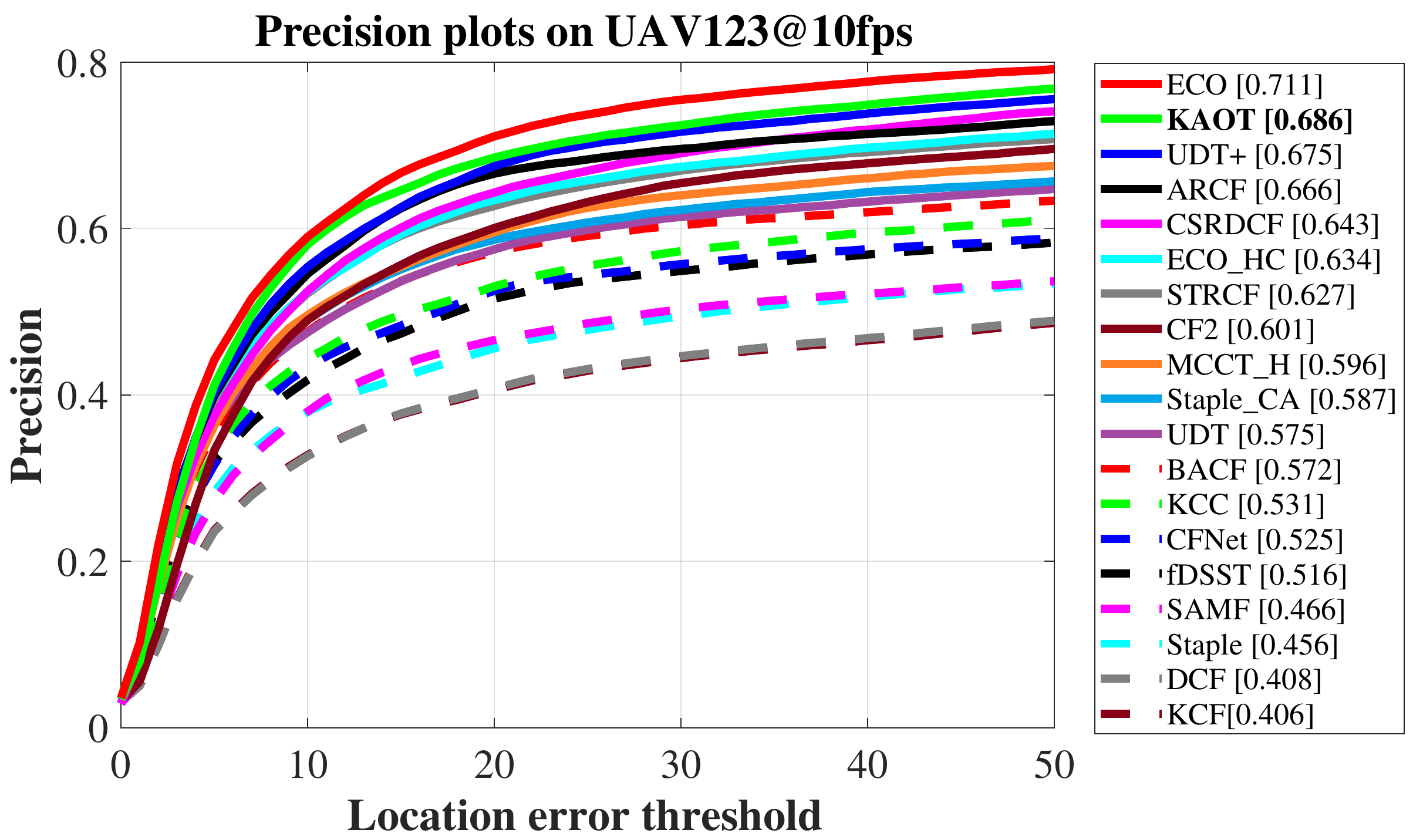}
			%\caption{fig2}
		\end{minipage}
	}%
	\subfigure{
		\begin{minipage}[t]{0.47\linewidth}
			\centering
			\includegraphics[width=3.1in]{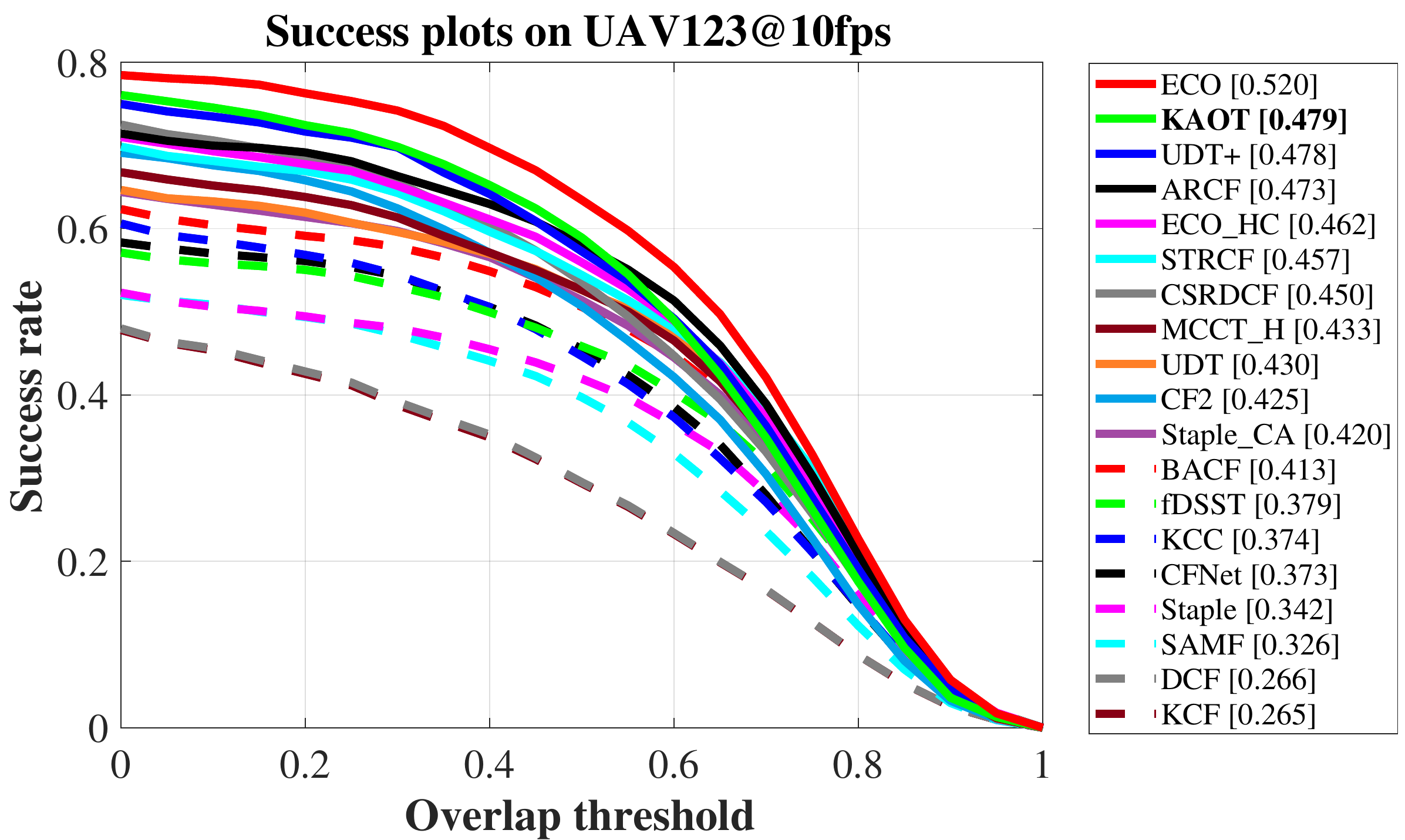}
			%\caption{fig2}
		\end{minipage}
	}%
	\centering
	\caption{\textbf{Precision and success plots} based on one-pass-evaluation \cite{Wu2015TPAMI} of KAOT and other real-time trackers on DTB70 \cite{li2017AAAI} and UAV123@10fps \cite{Mueller2016ECCV}.}
	\label{fig:overall}
\end{figure*}
Lagrangian parameter is updated as follows:
%%%%%%%%%%%%%%%%
\begin{equation}\label{Lagragian_format}
\small
	\mathbf{\hat{\zeta}} _ {j+1} = \mathbf{\hat{\zeta}} _ {j} + \mu \left(\mathbf{\hat{g}} ^ {*} _ {j+1} - \mathbf{\hat{w}} ^ {*} _ {j+1} \right) \ , 
\end{equation}
and $\mathbf{\hat{w}}^{*}_{j+1}$ is obtained through the following formula:
\begin{equation}
\small
	\mathbf{\hat{w}}^{*}_{j+1} = \left(\mathbf{I}_D \otimes \mathbf{FB}^{\top}\right) \mathbf{{w}}^{*}_{j+1} ,
\end{equation}
%%%%%%%%%%%%%%%%
\noindent subscript $j$ denotes the the value at last iteration and subscript $j+1$ denotes the value at current iteration. 
%%%%%%%%%%%%%%%%%%%%%%%%%%%%%%%%%%%%%%%%%%%%%%%%%%%%%%%%%%%%%%%%%%%%%%%%%%%%%%%
\subsection{Context patches scoring scheme}\label{subsec:scoring}
This work adopts a simple but effective scheme for measuring the score of context patches through Euclidean distance. Specifically, the size of omni-directional patches located around the object is the same as that of the object. The score of patch $p$ is calculated as follows:
%%%%%%%%%%%%%%%%%
\begin{equation}\label{Equ:score}
S_p = \frac{min \lbrace w,h \rbrace} {\left| O O_p \right| }s \ ,
\end{equation}
%%%%%%%%%%%%%%%%%
where $\left| O O_p \right| $ denotes the Euclidean distance between the object and context patch $p$ $(p=1, 2, ..., P)$ (between center points) and $s$ is the base score which is a constant number. $w$, $h$ are respectively the width and height of the object rectangle. Through Eq.~(\ref{Equ:score}) , the patch which is closer to object, obtains a higher score for stronger penalization. 
%%%%%%%%%%%%%%%%%%%%%%%%%%%%%%%%%%%%%%%%%%%%%%%%%%%%%%%%%%%%%%%%
\subsection{Keyfilter updating strategy}\label{subsec:keyfilter}
Starting from the first frame, the keyfilter is generated at a certain frequency using keyframes and current keyfilter refers to the latest trained keyfilter, as shown in Fig.~\ref{fig:mainstructure}. Current filter is restricted by current keyfilter through the punishment introduced by the gap between current filter and keyfilter. In other words, current keyfilter is updated every $c$ frames ($c=8$ in this work). When the $(n+1)$th keyframe arrives (frame $k=c  \times n + 1$), the filter of current frame (keyfilter $(n+1)$) is trained under influence from the keyfilter $n$. As for the non-keyframes after keyfilter $(n+1)$, the filters of them are learned with the restriction of current keyfilter (keyfilter $(n+1)$). The detailed work-flow of KAOT tracker is presented in Algorithm~\ref{alg:KAOTtrackerflow}.   
%%%%%%%%%%%%%%%%%%%%%%%%%%%%%%%%%%%%%%%%%%%%%%%%%
%%%%%%%%%%%%%%%%%%%%%%%%%%%%%%%%%%%%%%%%%%%%%%%%%%%%%%%%%%%%%%%%
%%%%%%%%%%%%%%%%%%%%% Section 4: EXPERIMENT %%%%%%%%%%%%%%%%%%%%
%%%%%%%%%%%%%%%%%%%%%%%%%%%%%%%%%%%%%%%%%%%%%%%%%%%%%%%%%%%%%%%%
\begin{algorithm}[!t]
	%\SetAlgoNoLine
	\caption{KAOT tracker}
	\label{alg:KAOTtrackerflow}
	\KwIn {\hspace{0.35cm}Location of tracked object on frame ${k-1}$, \\
		\hspace{1.45cm}Current keyfilter $\mathbf{ \tilde{w} }$, \\
		\hspace{1.45cm}Keyfilter updating $Stepsize$. }
	\KwOut {Location and scale of object on frame ${k}$}
	\For{$i=2$ to end}{
		Extract features from the region of interest (ROI) \\
		Convolute $\mathbf{\hat{g}}_{k-1}$ with $\mathbf{\hat{x}}^{i}_{\text{detect}}$ on different scales to generate response maps\\
		Find the peak position of map and output\\
		Update object model \\
		\eIf{k mod Stepsize$\times2==0$ }{Calculate $S_p$ $(p=1,2,...,8)$ by Eq.~(\ref{Equ:score}) \\Learn CF $\mathbf{w}_k$ by Eq.~(\ref{subproblem_w}), Eq.~(\ref{equ:subprobg}) and Eq.~(\ref{Lagragian_format}) \\$\mathbf{ \tilde{w} }=\mathbf{w}_k$}
		{\eIf{k mod Stepsize $==0$}{$S_p=0$ $(p=1,2,...,8)$\\Learn $\mathbf{w}_k$ by Eq.~(\ref{subproblem_w}), (\ref{equ:subprobg}) and Eq.~(\ref{Lagragian_format})\\$\mathbf{ \tilde{w} }=\mathbf{w}_k$}
			{$S_p=0$ $(p=1,2,...,8)$ \\ Learn $\mathbf{w}_k$ by Eq.~(\ref{subproblem_w}), Eq.~(\ref{equ:subprobg}) and Eq.~(\ref{Lagragian_format})}}
		{Start detection of next frame}		
	}
\end{algorithm}
\section{EXPERIMENTS}\label{sec:EXPERIMENT}
In this section, the presented KAOT tracker is rigorously evaluated on two difficult UAV datasets, i.e., DTB70~\cite{li2017AAAI} and UAV123@10ps~\cite{Mueller2016ECCV}, with overall 193 image sequences captured by drone camera.
% The challenging attributes of aerial tracking are classified systematically and covered exhaustively in the two datasets, so the tracker can be tested comprehensively from different aspects.
The tracking results are compared with the state-of-the-art trackers including both real-time (>=12 FPS) and non-real-time (< 12FPS) ones, i.e., ARCF~\cite{Huang2019ICCV}, UDT~\cite{Wang2019CVPR}, UDT+~\cite{Wang2019CVPR}, MCCT~\cite{Wang2018CVPR}, MCCT-H~\cite{Wang2018CVPR}, CSR-DCF~\cite{Luke2017CVPR}, STRCF~\cite{Li2018CVPR}, DeepSRTCF~\cite{Li2018CVPR}, ECO~\cite{Danelljan2017CVPR}, ECO-HC~\cite{Danelljan2017CVPR}, BACF~\cite{Galoogahi2017ICCV}, Staple~\cite{Bertinetto2016TPAMI}, Staple-CA~\cite{Mueller2017CVPR}, CF2~\cite{Ma2015ICCV},  DCF~\cite{Mueller2017CVPR}, DSST~\cite{Danelljan2017TPAMI}, KCF~\cite{Henriques2015TPAMI}, KCC~\cite{wang2018kernel}, SAMF~\cite{Yang2015ECCVW}, ADNet~\cite{Yun2017CVPR}, CFNet~\cite{Valmadre2017CVPR}, MCPF~\cite{Zhang2017CVPR}, IBCCF~\cite{Li2017ICCVW}. This work evaluates the trackers based on protocol in two datasets respectively~\cite{li2017AAAI,Mueller2016ECCV}. Noted that the real-time trackers are trackers with enough speed for UAV real-time applications. 
\subsection{Implementation details}
\label{subsec:EvaCri}
KAOT adopts both the hand-crafted and deep features, i.e., histogram oriented gradient (HOG)~\cite{Dalal2005CVPR}, color name (CN)~\cite{Danelljan2014CVPR} and conv3 layer from VGG-M network~\cite{Simonyan2015ICLR}. The value of $\gamma$ is set as 10, and the base score s is set as 0.28. ADMM iteration is set to 2 for raising efficiency. All trackers are implemented in MATLAB R2017a and all the experiments are conducted on the same computer with an i7-8700K processor (3.7GHz), 48GB RAM and NVIDIA GTX 2080 GPU. It is noted that the original codes without any modification are employed in this work for fair comparison.
\begin{figure*}[!t]
	\centering
	\subfigure{
		\begin{minipage}[t]{0.33\linewidth}
			\centering
			\includegraphics[width=2.2in]{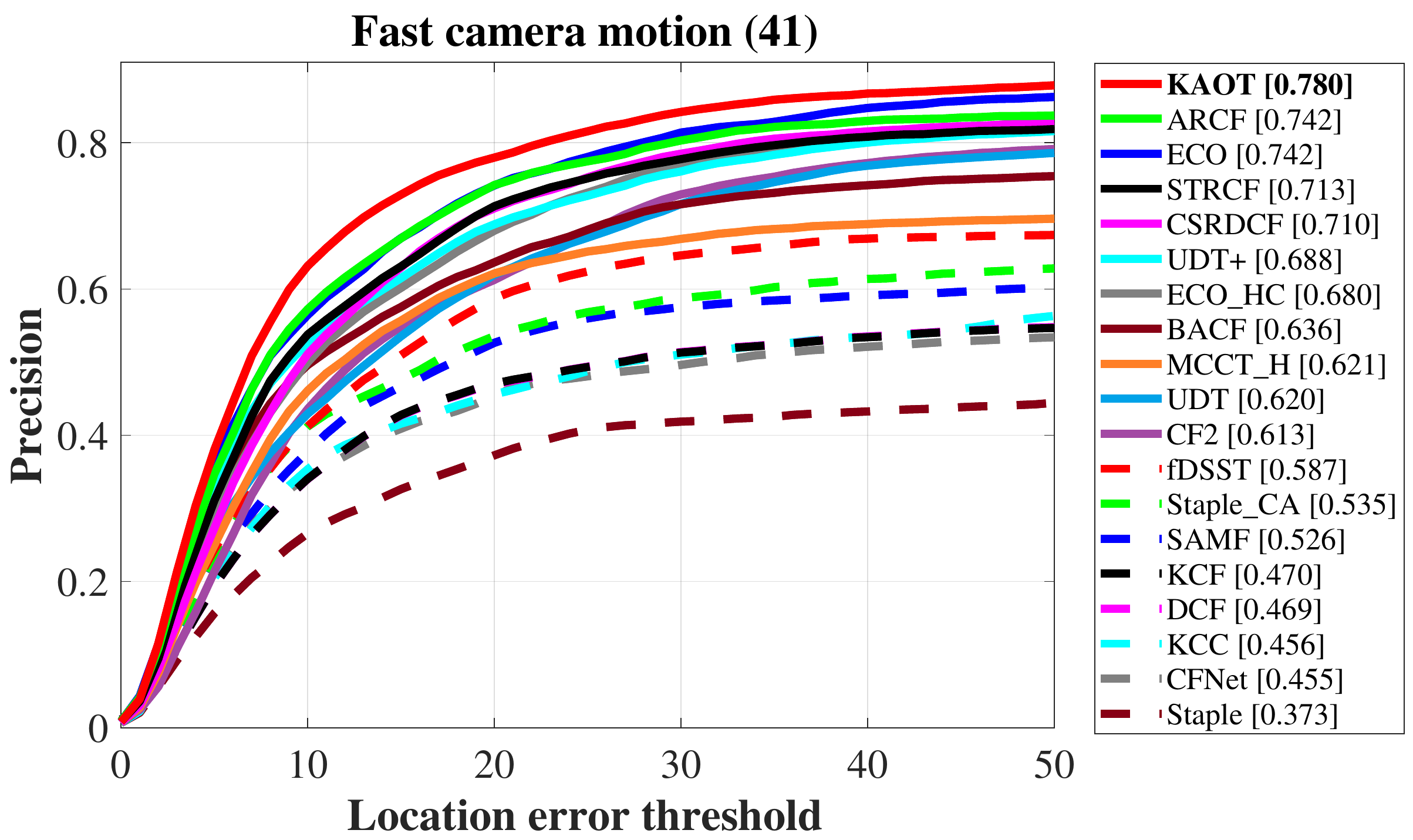}
			%\caption{fig1}
		\end{minipage}%
	}%
	\subfigure{
		\begin{minipage}[t]{0.33\linewidth}
			\centering
			\includegraphics[width=2.2in]{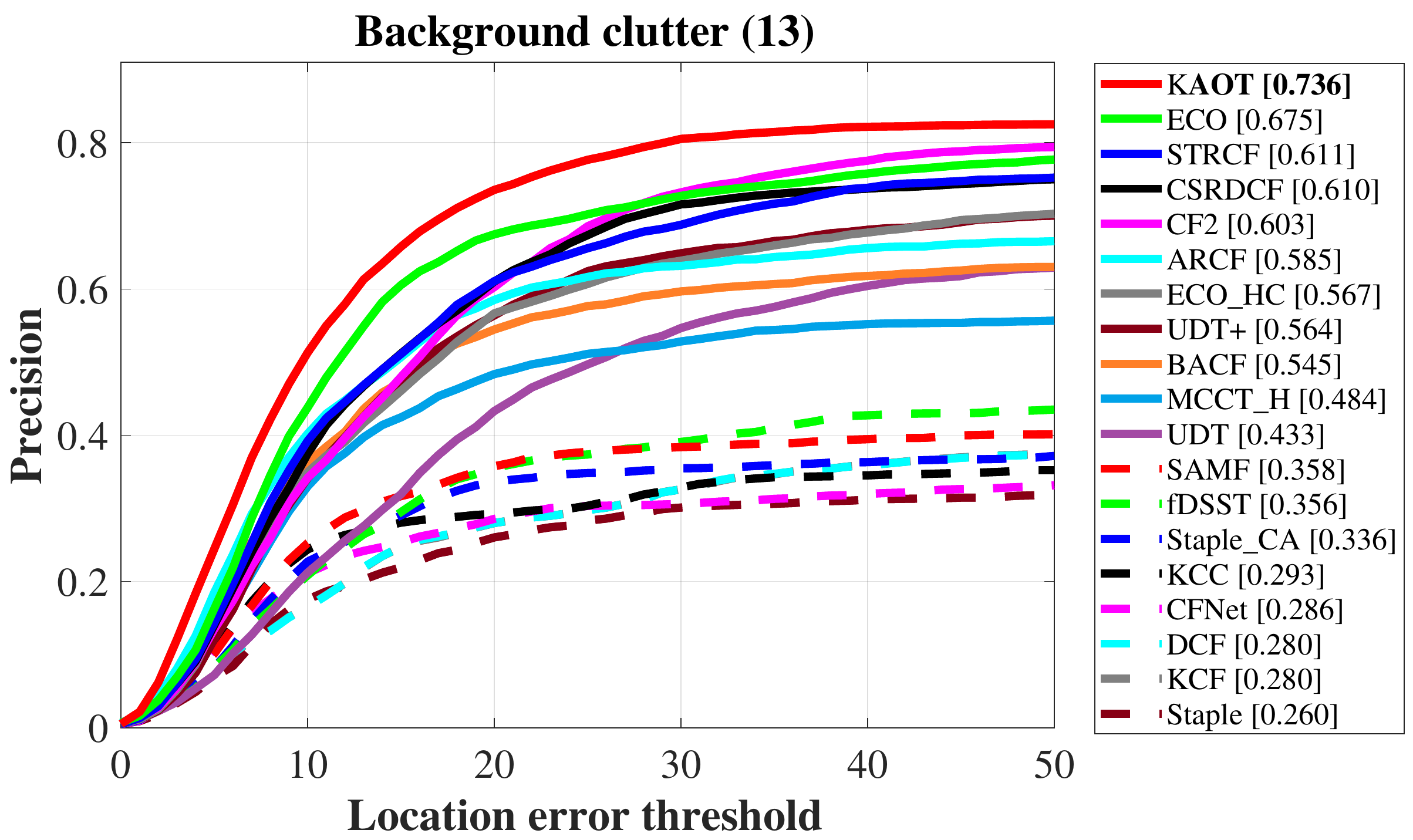}
			%\caption{fig2}
		\end{minipage}%
	}%
	\subfigure{
		\begin{minipage}[t]{0.33\linewidth}
			\centering 
			\includegraphics[width=2.2in]{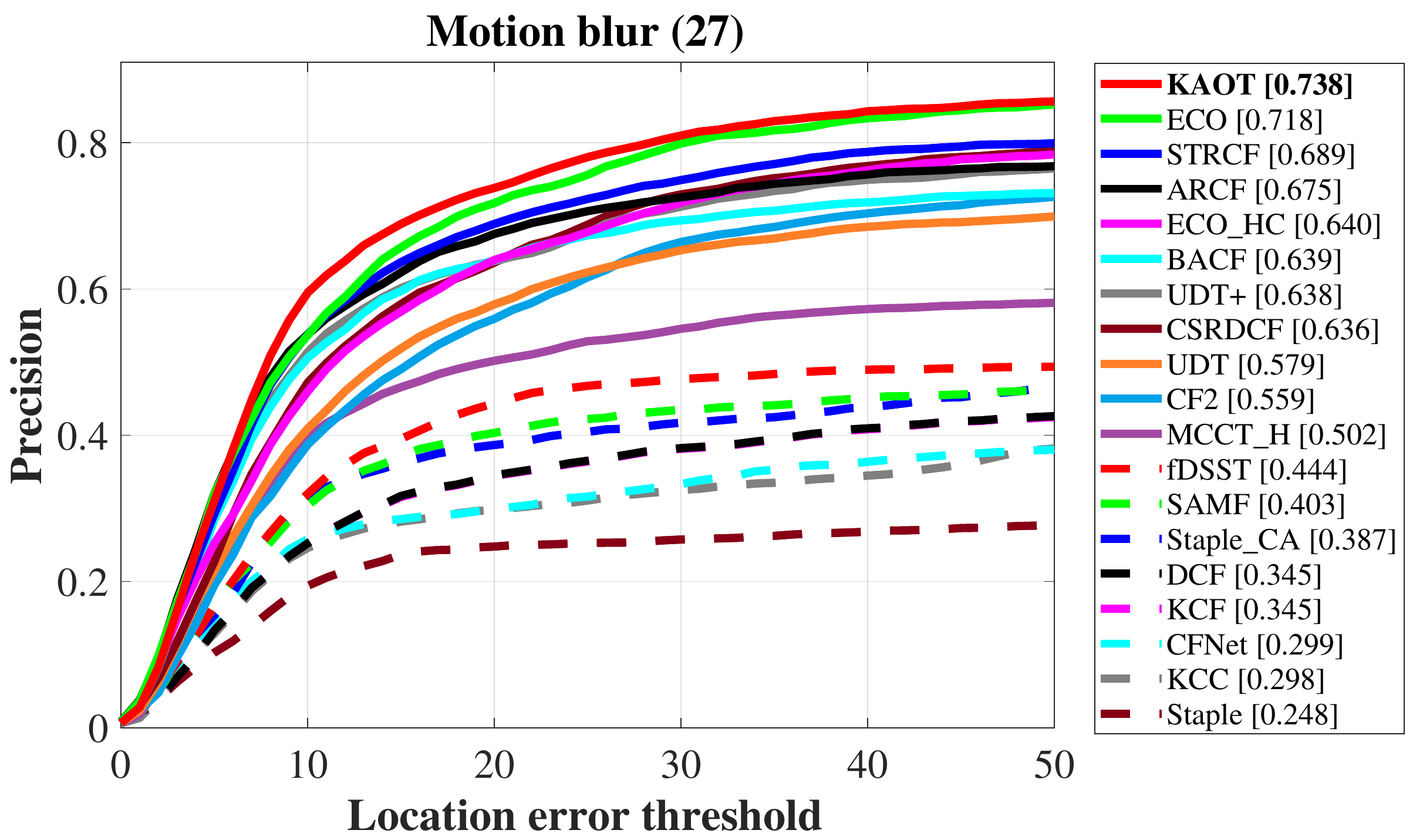}
			%\caption{fig2}
		\end{minipage}
	}%
	
	\subfigure{
		\begin{minipage}[t]{0.33\linewidth}
			\centering
			\includegraphics[width=2.2in]{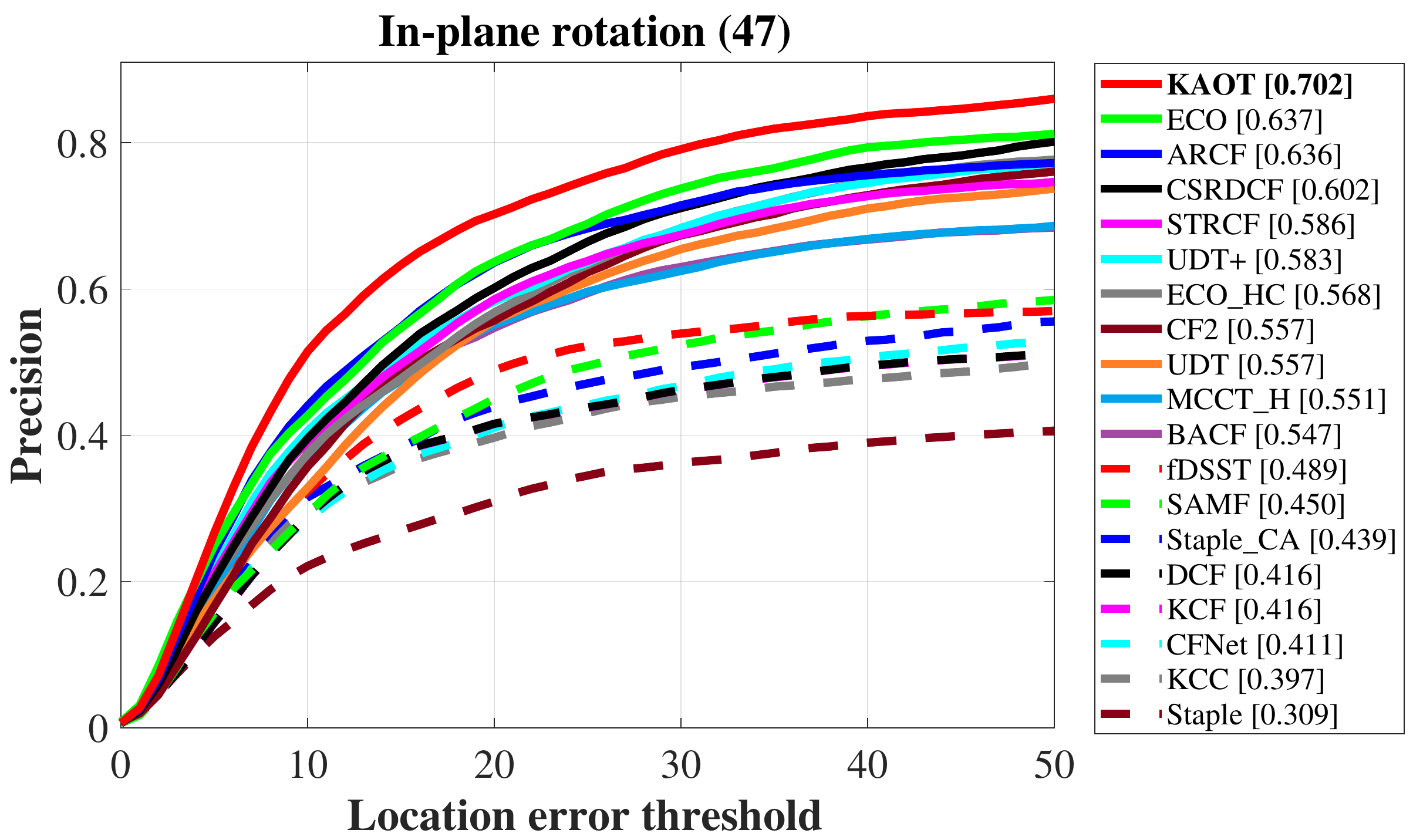}
			%\caption{fig1}
		\end{minipage}%
	}%
	\subfigure{
		\begin{minipage}[t]{0.33\linewidth}
			\centering
			\includegraphics[width=2.2in]{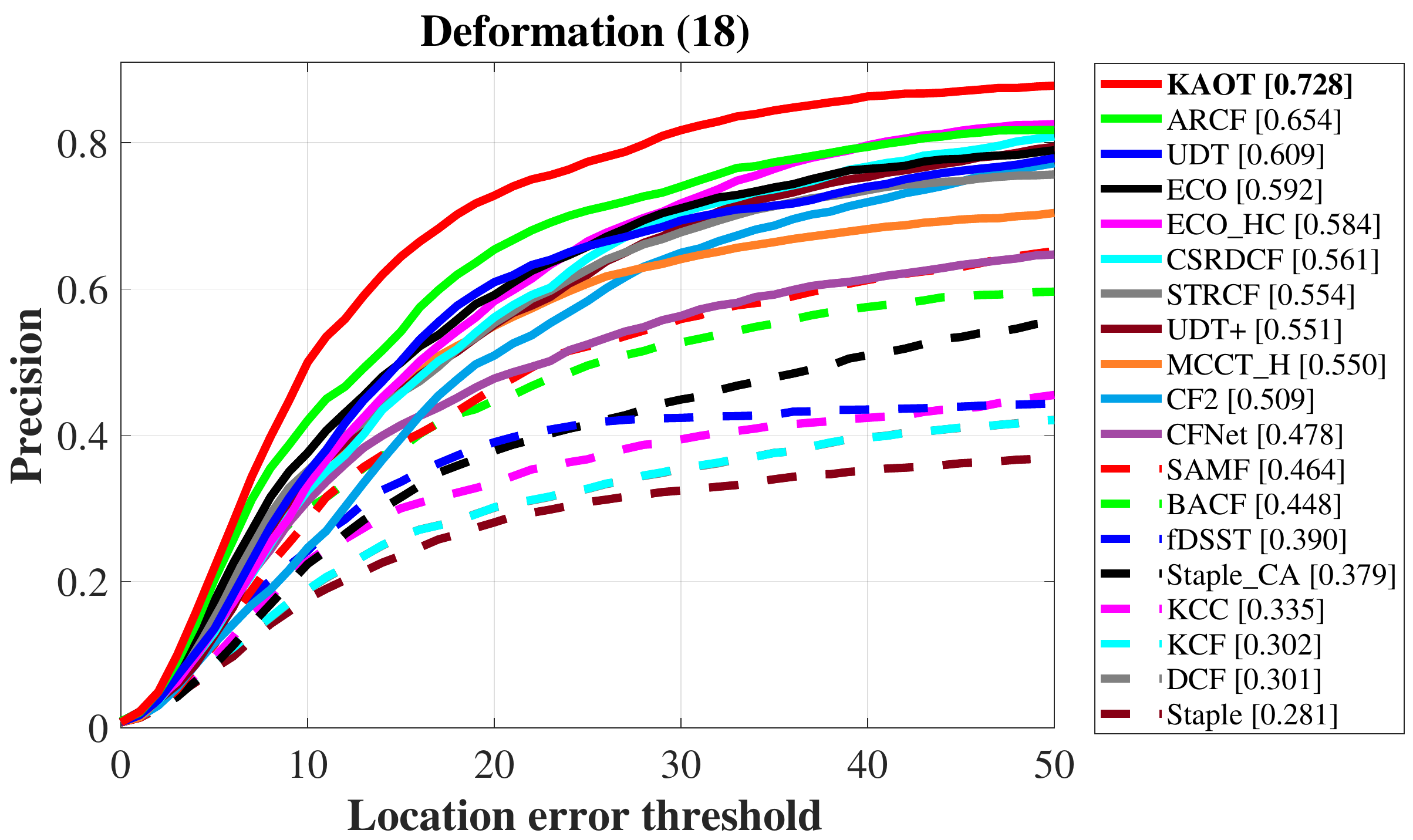}
			%\caption{fig1}
		\end{minipage}%
	}%
	\subfigure{
		\begin{minipage}[t]{0.33\linewidth}
			\centering
			\includegraphics[width=2.2in]{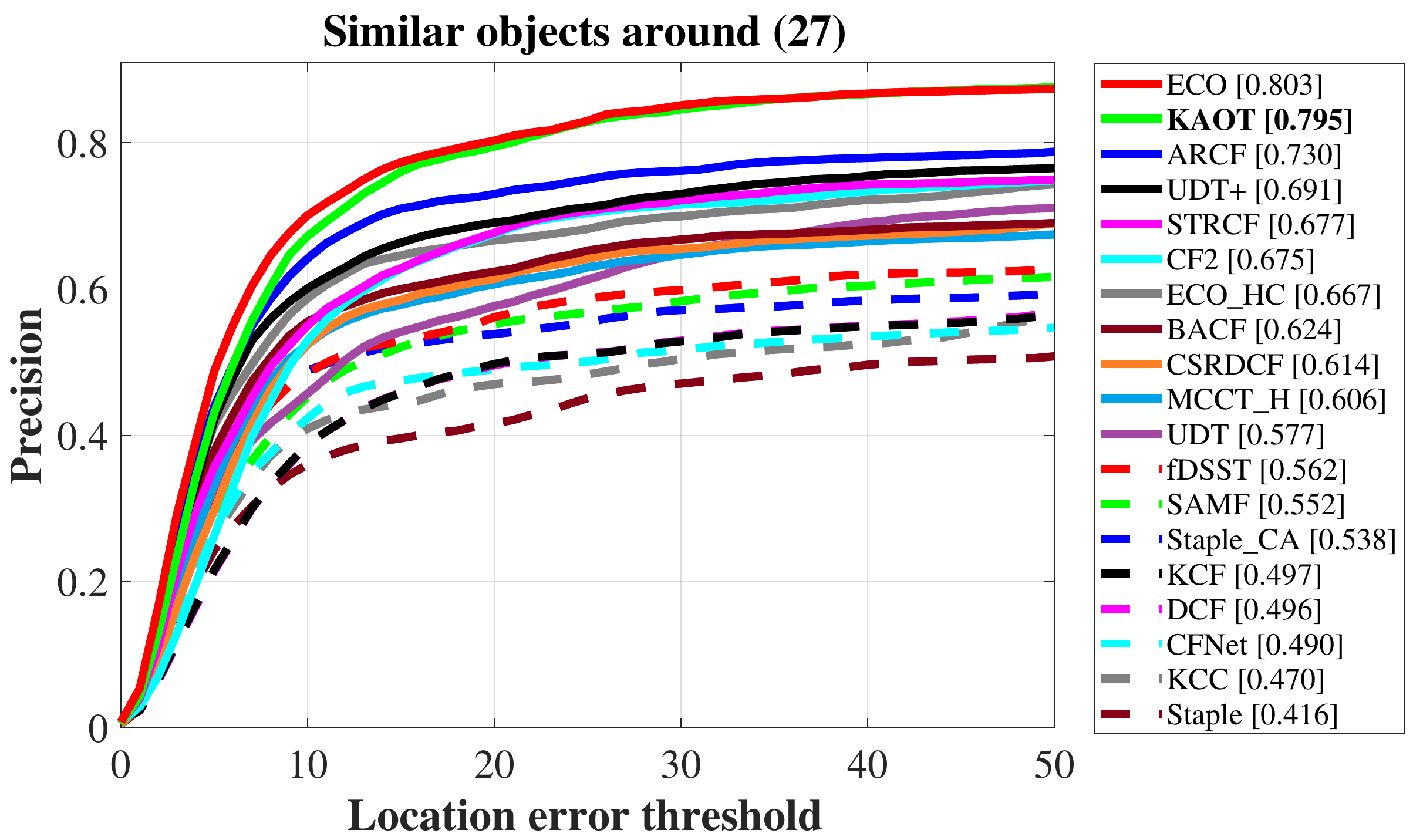}
			%\caption{fig1}
		\end{minipage}%
	}%
	\centering
	\caption{\textbf{Attribute based evaluation on precision.} KAOT ranks first place on five out of six challenging attributes.}
	\label{fig:attribute}
\end{figure*}
%%%%%%%%%%%
\begin{center}
	\begin{table*}[!t]
		\setlength{\tabcolsep}{1.5mm}
		\centering
		\caption{Average precision (threshold at 20 pixels) and Speed ((fps, * means GPU speed, otherwise CPU speed) ) of top ten real-time trackers. {\color{red}Red} , {\color{green}green}, and {\color{blue}blue} fonts respectively indicates the best, second, and third place in ten trackers. }
		\linespread{1.5}
		\footnotesize
		\begin{tabular}{ccccccccccc}
			\hline\hline
			&{\textbf{KAOT}}&{ECO}\cite{Danelljan2017CVPR}&{ARCF}\cite{Huang2019ICCV}&{UDT+}\cite{Wang2019CVPR}&{STRCF}\cite{Li2018CVPR}&{CSRDCF}\cite{Luke2017CVPR}&{ECO\_HC}\cite{Danelljan2017CVPR}&{CF2}\cite{Ma2015ICCV}&{MCCT\_H}\cite{Wang2018CVPR}&{UDT}\cite{Wang2019CVPR} \\\hline
			
			%			&&\textbf{}\cite{Danelljan2017CVPR}&\textbf{}\cite{Huang2019ICCV}&\textbf{}\cite{Wang2019CVPR}&\textbf{}\cite{Li2018CVPR}&\textbf{}\cite{Luke2017CVPR}&\textbf{}\cite{Danelljan2017CVPR}&\textbf{}\cite{Ma2015ICCV}&\textbf{}\cite{Wang2018CVPR}&\textbf{}\cite{Wang2019CVPR}& \\\hline\hline
			%			{When} 
			%			&  & 2017& 2019 &2019& 2019 & 2018 & 2017&2017 &2015& 2018 & \\
			%			{Where} 
			%			&  & CVPR& CVPR &ICCV& CVPR & CVPR & CVPR&CVPR &ICCV& CVPR & \\\hline\hline
			{{Avg. precision}} & \textcolor[rgb]{ 1,  0,  0}{72.2}  & \textcolor[rgb]{ 0,  1,  0}{71.7} & \textcolor[rgb]{ 0,  0,  1} {68.0}& {66.5} & {63.8} & 63.5& 64.3 &62.5&60.3&58.9  \\\hline
			{{Speed (FPS)}} 
			& {14.7}* & 11.6* & {15.3} & {43.4}* & {26.3} & 11.8& \textcolor[rgb]{ 1,  0,  0}{62.19} &14.4* &\textcolor[rgb]{ 0,  1,  0}{59.0} &\textcolor[rgb]{ 0,  0,  1}{57.5}* \\\hline\hline	
		\end{tabular}%
		\label{tab:fps}%
	\end{table*}%
\end{center}
\subsection{Comparison with real-time trackers}
\label{subsec:overall}
\subsubsection{Overall performance}
Figure~\ref{fig:overall} demonstrates the overall performance of KAOT with other state-of-the-art real-time trackers on DTB70 and UAV123@10fps. On DTB70 dataset, KAOT
(0.757) has an advantage of 4.4\% and 9.1\% over the second and third best tracker ECO (0.722), ARCF (0.694) respectively in precision, along with a gain of 0.2\% and 6.6\% over the second (ECO, 0.502) and third best tracker (ARCF, 0.472) respectively in AUC. On UAV123@10fps dataset, KAOT (0.686, 0.479) ranks second place followed by the third place UDT+ (0.675, 0.478). ECO is the only tracker performing better than KAOT. Nevertheless, it utilizes continuous operator to fuse the feature maps elaborately, while KAOT just uses the simple BACF as baseline. Notice that ECO can further enhance its performance with our framework. Average precision on the two datasets and speed (evaluated on DTB70) are reported in Table~\ref{tab:fps}. KAOT is 27\% faster than ECO when achieving higher precision.

\noindent \textbf{Discussions:} DTB70~\cite{li2017AAAI} dataset is recorded on a drone with more frequent and drastic displacements  compared to UAV123@10fps~\cite{Mueller2016ECCV}, thus increasing the tracking difficulties. Our method exhibits relatively big advantages on DTB70, proving the robustness of our method in the scenarios of strong motion.
\begin{figure}[!t]
	\centering
	\includegraphics[width=0.48\textwidth]{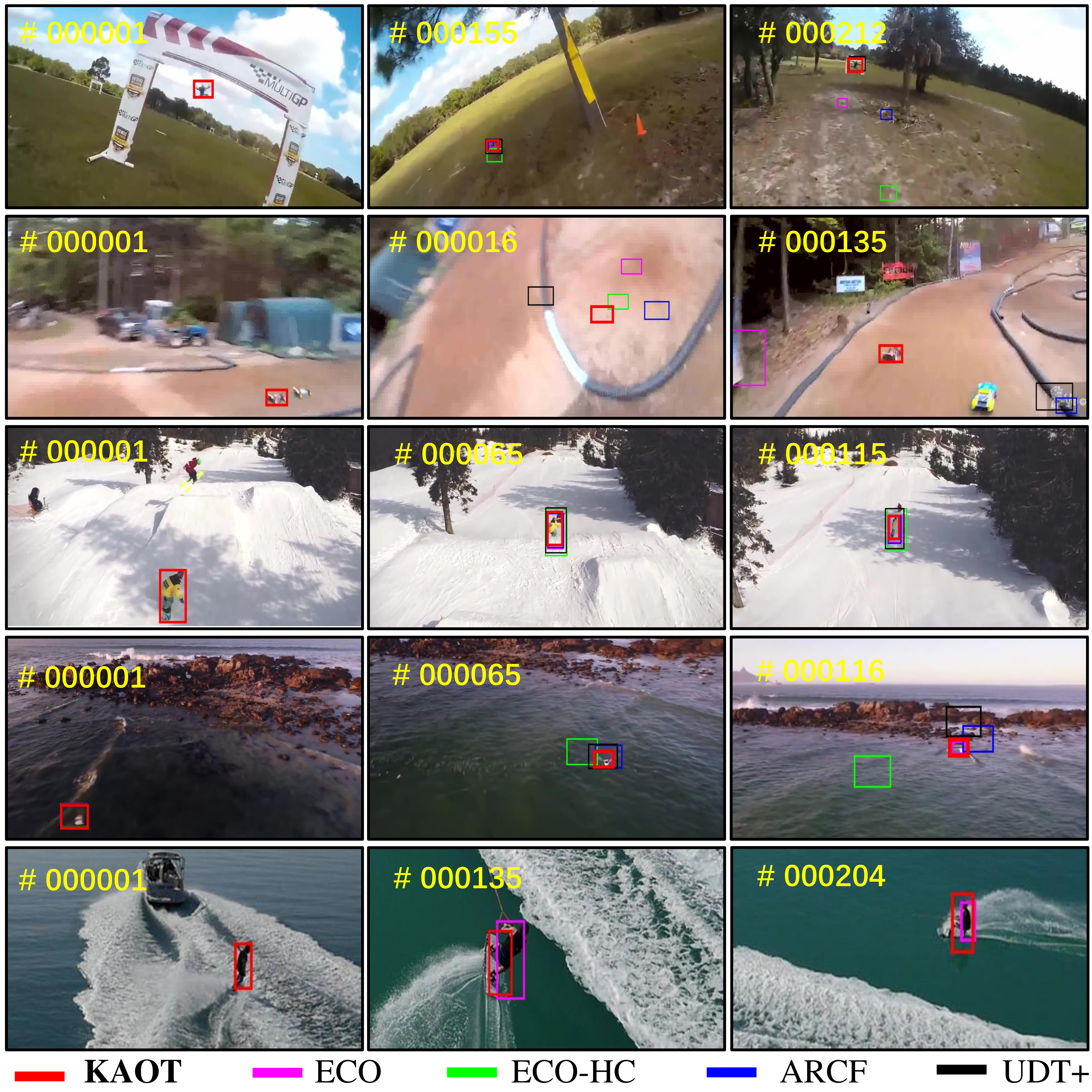}
	\caption{\textbf{Qualitative evaluation.} From the top to bottom is respectively the sequence $ChasingDrones$, $RcCar6$, $SnowBoarding2$, $Gull1$ and $wakeboard2$. Code and UAV tracking video are: \url{https://github.com/vision4robotics/KAOT-tracker} and \url{https://youtu.be/jMfmHVRqv3Y}.}
	\label{fig:comp}
\end{figure}
\subsubsection{Attribute-based performance}
\label{subsec:attr}
Precision plots of six challenging attributes are demonstrated in Figure~\ref{fig:attribute}. In the cases of background clutter, KAOT improves the ECO by 9.0\% in light of the intermittent context learning which can suppress the background distraction effectively. In situations of in-plane rotation and deformation, KAOT has a superiority of  10.2\% and 23.0\% respectively compared to ECO. This is attributed to the keyfilter restriction, which can prevent the filter from aberrant variation.  In addition, KAOT exhibits excellent performance in the scenario of fast camera motion and motion blur, which is desirable in aerial tracking.
%%%%%%%%%%%
%%%%%%%%%%%
\subsubsection{Qualitative evaluation}
\begin{figure}[t]
	\centering
	\subfigure{
		\begin{minipage}[t]{1\linewidth}
			\centering
			\includegraphics[width=3.3in]{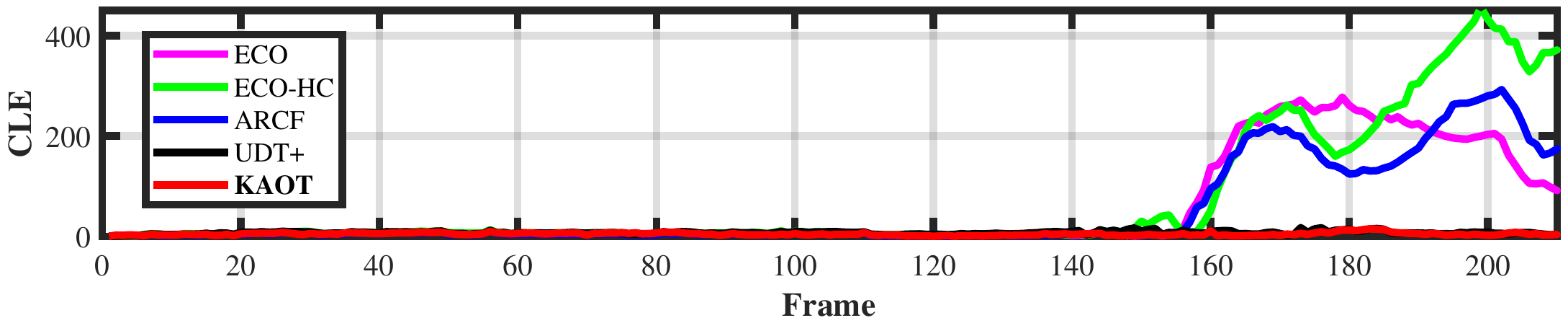}
			%\caption{fig1}
		\end{minipage}%
	}%
	
	\subfigure{
		\begin{minipage}[t]{1\linewidth}
			\centering
			\includegraphics[width=3.3in]{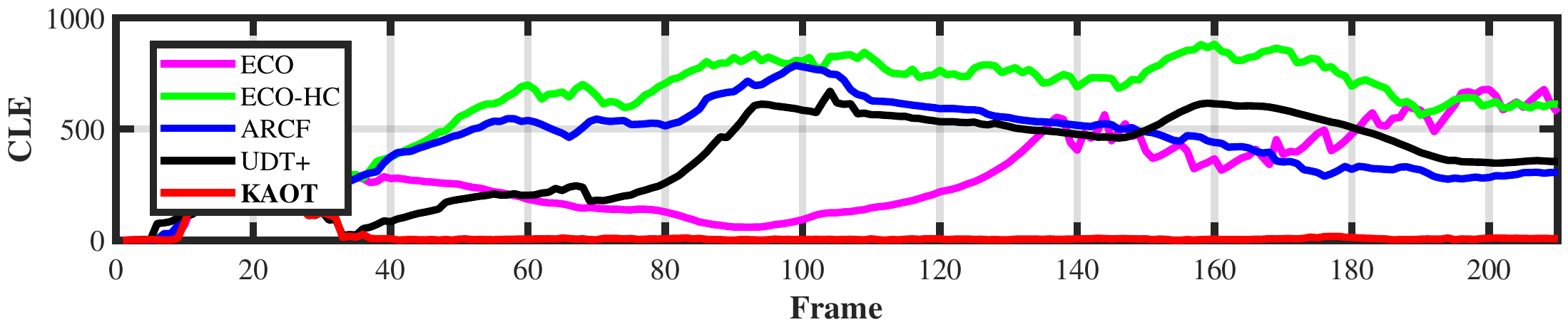}
			%\caption{fig1}
		\end{minipage}%
	}%
	
	\subfigure{
		\begin{minipage}[t]{1\linewidth}
			\centering
			\includegraphics[width=3.3in]{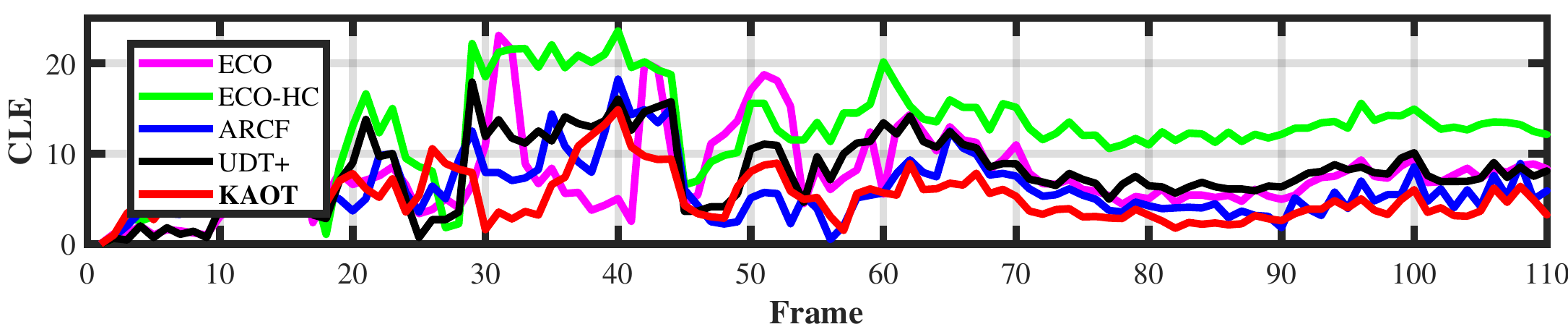}
			%\caption{fig1}
		\end{minipage}%
	}%
	
	\subfigure{
		\begin{minipage}[t]{1\linewidth}
			\centering
			\includegraphics[width=3.3in]{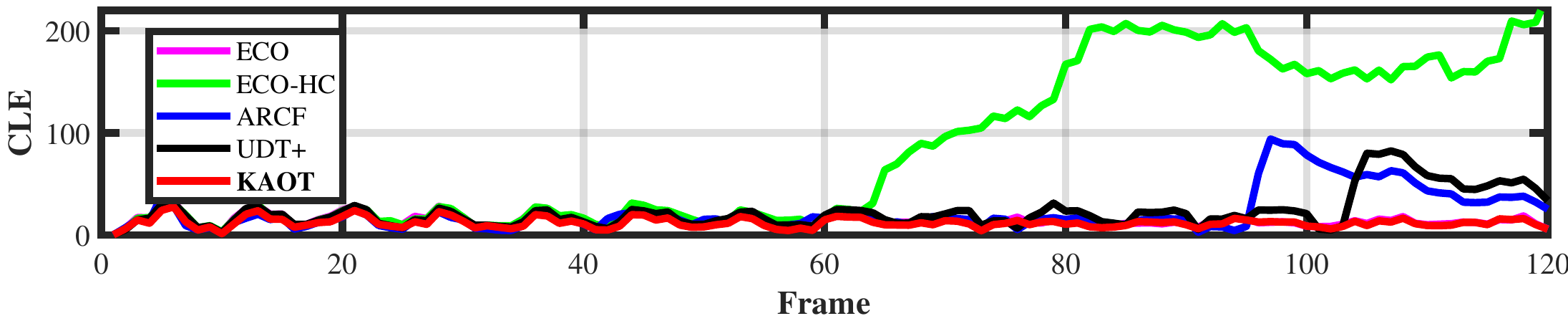}
			%\caption{fig1}
		\end{minipage}%
	}%
	
	\subfigure{
		\begin{minipage}[t]{1\linewidth}
			\centering
			\includegraphics[width=3.3in]{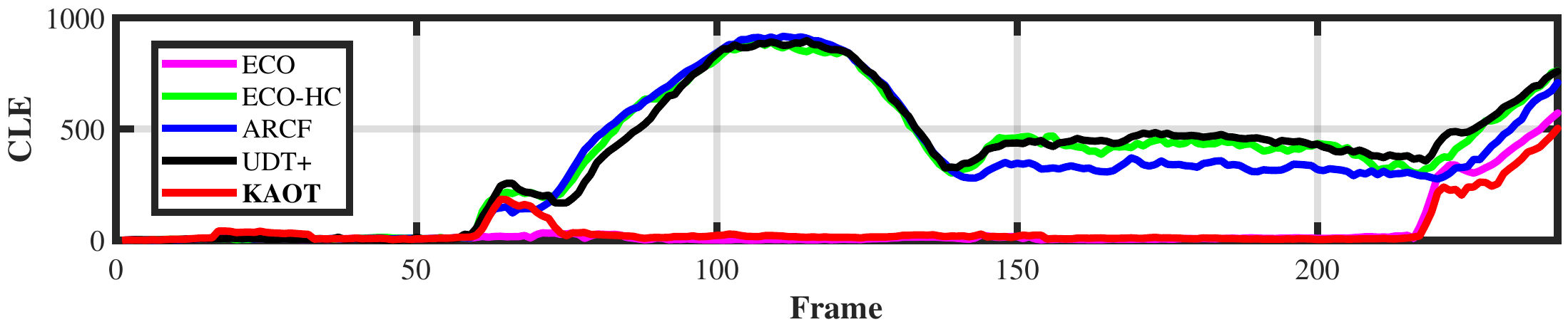}
			%\caption{fig1}
		\end{minipage}%
	}%
	\centering	
	\caption{\textbf{Illustration of CLE variations.} From top to bottom is the result from sequence $ChasingDrones$, $RcCar6$, $SnowBoarding2$, $Gull1$ and $wakeboard2$, respectively.}
	\label{fig:cle}
\end{figure}

Qualitative tracking results on five difficult UAV image sequences are shown in Figure~\ref{fig:comp}. Besides, the respective center location error (CLE) variations of five sequences are visualized in Figure~\ref{fig:cle}.
Specifically, in $ChasingDrones$ sequence where tracking is bothered by strong UAV motion, KAOT has effectively repressed the distraction of the context, so it can perform well despite the large movement in a certain complex context. Only the pre-trained UDT+ tracks successfully in addition to KAOT. Motion blur occurs in sequences $RcCar6$ and $Gull1$ (severe example is shown at frame 16 in $RcCar6$). In this situation, KAOT has kept tracking owing to the mitigated filter corruption. As for the last two sequences, keyfilter restriction and intermittent context learning have collaboratively contributed to successful tracking.

\subsection{Comparison with non-real-time trackers}
KAOT is also compared with five non-real-time trackers using deep neural network, as shown in Table~\ref{tab:deep}. To sum up, KAOT has the best performance in terms of both precision and speed on two benchmarks. In addition, compared to DeepSTRCF (using the same features as KAOT), our tracker has more robust performance in precision on both two datasets and is around 2.4 times faster than it. Therefore, the efficiency and accuracy of KAOT tracker can be proven.

\begin{center}
	\begin{table}[!b]
		\caption{Precision, success rate (the area under the curve), and Fps of KAOT as well as five non-real-time trackers. {\color{red}Red} , {\color{green}green}, and {\color{blue}blue} fonts respectively indicates the best, second, and third  performance.}
		\linespread{1.5}
		\footnotesize
		\begin{center}
			\begin{tabular}{c|c|c|c|c|c}
				\hline\hline
				&\multicolumn{2}{c|}{{DTB70}}&\multicolumn{2}{c|}{{UAV123@10fps}}&\multirow{2}{*}{FPS}\\\cline{1-5}
				{Trackers}&Prec.&AUC&Prec.&AUC&\\\hline
				%C-COT \cite{Danelljan2016ECCV}& \textcolor[rgb]{ 1,  0,  0}{76.9} & \textcolor[rgb]{ 1,  0,  0}{51.7} & \textcolor[rgb]{ 1,  0,  0}{70.6}& \textcolor[rgb]{ 0,  0,  0}{50.3} &1.10*\\
				MCPF\cite{Zhang2017CVPR}&{66.4} & {43.3}& 66.5 & {44.5}  &0.57* \\
				
				MCCT\cite{Wang2018CVPR}& \textcolor[rgb]{ 0,  0,  1}{72.5} & \textcolor[rgb]{ 0,  0,  1}{48.4}& \textcolor[rgb]{ 0,  1,  0}{68.4} & \textcolor[rgb]{ 0,  1,  0}{49.2} &\textcolor[rgb]{ 0,  1,  0}{8.49}*\\
				DeepSTRCF\cite{Li2018CVPR}&\textcolor[rgb]{ 0,  1,  0}{73.4} & \textcolor[rgb]{ 1,  0,  0}{50.6} & \textcolor[rgb]{ 0,  0,  1}{68.2} & \textcolor[rgb]{ 1,  0,  0}{49.9}  &\textcolor[rgb]{ 0,  0,  1}{6.18}*\\
				
				IBCCF\cite{Li2017ICCVW}& {66.9} &46.0 & {65.1} & \textcolor[rgb]{ 0,  0,  1}{48.1}  &2.28*\\
				ADNet \cite{Yun2017CVPR}& {63.7} & 42.2 & {62.5} & 43.9  &6.87*\\
				\textbf{KAOT} & \textcolor[rgb]{ 1,  0,  0}{75.7} & \textcolor[rgb]{ 0,  1,  0}{50.3} & \textcolor[rgb]{ 1,  0,  0}{68.6} &\textcolor[rgb]{ 0,  0,  0}{47.9}  &\textcolor[rgb]{ 1,  0,  0}{14.69}*\\\hline\hline
				
			\end{tabular}%
		\end{center}
		\label{tab:deep}%
	\end{table}%
\end{center}
%%%%%%%%%%%
\subsection{Limitations and future works}
\label{subsec:limitations and future work}
\noindent \textbf{Keyframe selection:}
This work only adopts a simple periodic keyframe selection mechanism, which is possible to introduce distraction when the tracking on the keyframes is not reliable. More elaborated strategy can be employed to adaptively choose the keyframe and further enhance the robustness.

\noindent \textbf{Re-detection and rotation:}
Though KAOT performs favorably in the situations of drastic appearance change like blur, deformation, etc., it is still limited when the object disappear for a long time. Also, KAOT can not handle the rotation situations. Thus the re-detection and rotation-aware modules can be added to raise the performance.

\noindent \textbf{Speed:}
The speed of KAOT is around 15 fps with a GPU and can be used in real-time applications. However, KAOT tracker is implemented on MATLAB platform and the code is not optimized, so the speed can be further improved.

%%%%%%%%%%%%%%%%%%%%%%%%%%%%%%%%%%%%%%%%%%%%%%%%%%%%%%%%%%%%%%%%
%%%%%%%%%%%%%%%%%%%%% Section 5: CONCLUSIONS %%%%%%%%%%%%%%%%%%%
%%%%%%%%%%%%%%%%%%%%%%%%%%%%%%%%%%%%%%%%%%%%%%%%%%%%%%%%%%%%%%%%

\section{CONCLUSIONS}\label{sec:CONCLUSIONS}
This work proposes keyfilter-aware object tracker to repress the filter corruption and lower the redundancy of context learning. Extensive experiments on two authoritative datasets have validated our tracker performs favorably in precision, with enough speed for real-time applications. This keyfilter-aware framework and intermittent context learning strategy can also be used in other trackers like C-COT~\cite{Danelljan2016ECCV} and STRCF~\cite{Li2018CVPR} to further boost their performance. We strongly believe that our method can be used in practice and promote the development of UAV tracking applications.
\\
%%%%%%%%%%%%%%%%%%%%%%%%%%%%%%%%%%%%%%%%%%%%%%%%%%%%%%%%%%%%%%%%%%%%%%%%%%%%%%%%
%%%%%%%%%%%%%%%%%%%%%%%%%%%%%%%%%%%%%%%%%%%%%%%%%%%%%%%%%%%%%%%%%%%%%%%%%%%%%%%%
%%%%%%%%%%%%%%%%%%%%%%%%%%%%%%%%%%%%%%%%%%%%%%%%%%%%%%%%%%%%%%%%%%%%%%%%%%%%%%%%
%%%%%%%%%%%%%%%%%%%%%%%%%%%%%%%%%%%%%%%%%%%%%%%%%%%%%%%%%%%%%%%%
%%%%%%%%%%%%%%%%%%%%%  ACKNOWLEDGMENT %%%%%%%%%%%%%%%%%%%
%%%%%%%%%%%%%%%%%%%%%%%%%%%%%%%%%%%%%%%%%%%%%%%%%%%%%%%%%%%%%%%%
%%%%%%%%%%%
\section*{ACKNOWLEDGMENT}
This work is supported by the National Natural Science Foundation of China (No. 61806148) and the Fundamental Research Funds for the Central Universities (No. 22120180009).

%%%%%%%%%%%%%%%%%%%%%%%%%%%%%%%%%%%%%%%%%%%%%%%%%%%%%%%%%%%%%%%%%%%%%%%%%%%%%%%%

\bibliographystyle{IEEEtran}  %这是你要使用的格式,比如要投IEEE,就写IEEEtran
\bibliography{IEEEabrv,ref}%这个是加载你的bib,你可以理解从文献数据库中加载要引用的文献

\end{document}